\documentclass{article}

\usepackage{hyperref}
\usepackage{bookmark}
\usepackage{multicol}
\usepackage{graphicx}
\usepackage{amsfonts}
\usepackage{amsmath}
\usepackage{natbib}

\begin{document}
\title{Crowdsourcing for Beyond Polarity Sentiment Analysis\\A Pure Emotion Lexicon}

\author{
  Giannis Haralabopoulos\\
  University of Southampton\\
  \texttt{Giannis.Haralabopoulos@soton.ac.uk}
  \and
  Elena Simperl\\
  University of Southampton\\
  \texttt{E.Simperl@soton.ac.uk}
}

\maketitle

\begin{abstract}
Sentiment analysis aims to uncover emotions conveyed through information. In its simplest form, it is performed on a polarity basis, where the goal is to classify information with positive or negative emotion. Recent research has explored more nuanced ways to capture emotions that go beyond polarity. For these methods to work, they require a critical resource: a lexicon that is appropriate for the task at hand, in terms of the range of emotions it captures diversity. In the past, sentiment analysis lexicons have been created by experts, such as linguists and behavioural scientists, with strict rules. Lexicon evaluation was also performed by experts or gold standards. In our paper, we propose a crowdsourcing method for lexicon acquisition, which is scalable, cost-effective, and doesn't require experts or gold standards. We also compare crowd and expert evaluations of the lexicon, to assess the overall lexicon quality, and the evaluation capabilities of the crowd.	
\end{abstract}

\section{Introduction}
\begin{sloppypar}

\begin{figure}
\begin{center}
\includegraphics[scale=0.5]{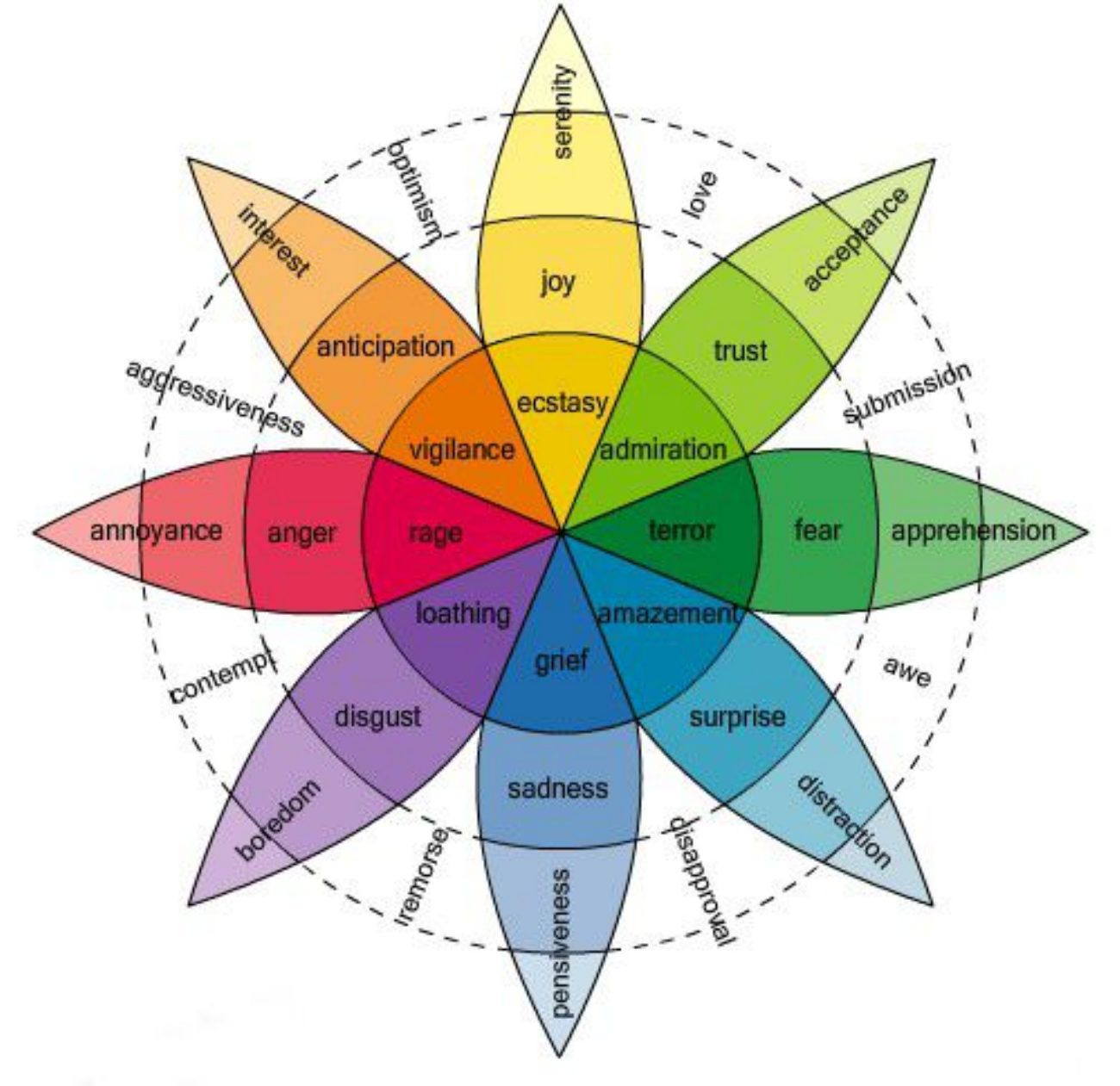}
\caption{The circumplex of emotions}
\label{fig:circumplex}
\end{center}
\end{figure}

Sentiment analysis aims to uncover the emotion conveyed through information. In online social networks, sentiment analysis is mainly performed for political and marketing purposes, product acceptance and feedback systems.  This involves the analysis of various social media information types, such as text \cite{khan2015combining}, emoticons and hashtags, or multimedia \cite{poria2016fusing}. However, to perform sentiment analysis, information has to be labelled with a sentiment. This relationship is defined in a lexicon.

Lexicon acquisition is a requirement for sentiment classification. During the acquisition process, individual or grouped information elements are labelled based on a class, usually an emotion. Sentiment classification is the task that uses the acquired lexicon and a classification method to classify a sentence, phrase, or social media submission as a whole, based on the aggregation of its labels. Thus, lexicon quality directly affects sentiment classification accuracy.

Both tasks can either be performed automatically \cite{erdmann2014feature} or manually \cite{iyer2015sentiment} where the labelling by linguists or researchers themselves \cite{abdul2014samar}. Apart from experts, manual labbeling can also be performed with the help of a wide network of people, known as crowdsourcing \cite{hu2013exploiting}. Crowdsourcing is widely used for polarity lexicons, but rarely for beyond polarity and never for the discovery of linguistic elements.

Sentiment analysis is commonly performed in polarity basis, i.e. the distinction between positive and negative emotion . These poles correspond to agreement and disagreement, or acceptance and disapproval, for candidates and products repsectively \cite{zhou2015latent}.

Beyond polarity (also known as pure emotion) sentiment analysis aims to uncover an exact emotion, based on emotional theories \cite{plutchik1980general,ekman1969pan}.  Applications such as sentiment tracking, marketing, text correction, and text to speech systems can be improved with the use of distinct emotion lexicons.

However, beyond polarity studies acquire lexicons based on a set of strict rules, and the evaluation of experts. These lexicons use only a single emotion per term \cite{bandhakavi2017lexicon}. The problems of these approaches is the lack of uniformity and contribution freedom when relying on gold standards, and high costs with low scalability when employing experts.  Natural Language Processing (NLP) applications that only rely on experts are less comprehensive, restricted, and not scalable, compared to crowdsourced NLP applications \cite{gabrilovich2007computing}.

This paper presents our approach for the acquisition of a multiclass and scalable crowdsourced pure emotion lexicon (PEL), based on Plutchik's eight basic emotions. Furthermore, the crowd is also responsible for identifying linguistic elements, namely intensifiers, negators, and stop words. Usually these elements are pooled from existing lists \cite{kiritchenko2016effect} created by experts. We also introduce a worker filtering method to identify and exclude dishonest or spamming contributors, that doesn't require gold standards. Our goal is to maintain an end to end automated work-flow for a crowdsourced (annotation and evaluation wise) lexicon acquisition process. Therefore, to highlight crowd's performance on evaluation, we compare evaluations from linguistic experts and the crowd itself.

\section{Related Work}

According to \cite{cabanac2002emotion}, an emotion is defined with reference to a list. Ekam et al. \cite{ekman1969pan} proposed the six basic emotions joy, anger, fear, sadness, disgust, and surprise. Years later, Plutchik \cite{plutchik1980general} proposed the addition of trust and anticipation as basic emotions, and presented a circumplex model of emotions as seen in Figure~\ref{fig:circumplex}, which defines emotional contradictions and some of the possible combinations.

Sentiment analysis aims to classify information based on the emotion conveyed. Depending on the number of classes/emotions required, we can separate the analysis into: polarity and beyond polarity.

Polarity sentiment analysis studies define two opposite emotional states, positive and negative, or good and bad, with the addition of a neutral state. Furthermore, some researchers have classified information on levels for each pole(e.g. very positive, positive, neutral, negative, very negative etc.), also known as fine grained sentiment analysis \cite{guzman2014users}.

Beyond polarity, also known as pure emotion, sentiment analysis is a more refined approach to the same problem with a wider range of possible emotion classes, see Figure~\ref{fig:circumplex}. Essentially, any sentiment analysis that involves specific emotional labelling, is considered as a beyond polarity analysis. Examples of emotional labels might be -but are not limited to-: sadness, boredom, joy, sadness, surprise, anger, fear, disgust etc.

As discussed in Section 1, one of the core tasks of sentiment analysis is lexicon acquisition. A lexicon can be acquired through manual or automatic annotation. However, natural language has a very subjective nature \cite{alm2011subjective} which significantly inhibits automated sentiment lexicon aqcuisition methods from achieving relevance equal to manual methods \cite{lu2011automatic}. Thus a lot of researchers choose to manually annotate their term corpora \cite{poria2015deep}, or use established lexicon such as WordNet, SentiWordNet, and various other lexicons \cite{guzman2014users}. Other studies combine manual labeling or machine learning with lexicons \cite{ortigosa2014sentiment}.

Manual lexicon acquisition is constrained by the number of people contributing to the task, and the number of annotations from each participant. These constraints can be eliminated by increasing the number of people involved, for instance, by using crowdsourcing \cite{brabham2008crowdsourcing}. Amazon's Mechanical Turk (MTurk)\footnote{https://www.mturk.com/} is a crowdsourcing platform frequently used for polarity sentiment lexicon acquisition via crowdsourcing \cite{nakov2016semeval}. MTurk is also used, for the annotation of one thousand tweets in \cite{davidov2010enhanced}, ten thousand terms in \cite{mohammad2013nrc} with gold standards, and the annotation of ninety five emoticons out of one thousand total emoticons found in \cite{zhao2012moodlens}. While \cite{poria2013enhanced} had one thousand four hundred terms labelled with a supervised machine learning and crowd validators. The challenge is to introduce a work-flow that is scalable, unsupervised and applicable to different information types.

The second core part in sentiment analysis, is sentiment classification. A classification that occurs at phrase/sentence/submission level, and is usually based on the aggregation of the term's labeled emotions. As with lexicon aqcuisition, the classification task can be automated \cite{guzman2014users} or performed manually \cite{borromeo2015automatic}.

Regardless of manual or automated sentiment classification, on textual information scenarios, term and phrase sentiment is the main input of the classification method. In some cases the decision might be totally different from the individual term emotion, leading to relabeling of the terms themselves \cite{socher2013recursive}. Manually labelled classification can achieve high relevance, but it requires additional resources, and is not easily scalable. On the other hand, automated processes are scalable but with lower relevance \cite{borromeo2015automatic}.

\section{Our approach}

\begin{figure}[ht]
\includegraphics[width=1\textwidth]{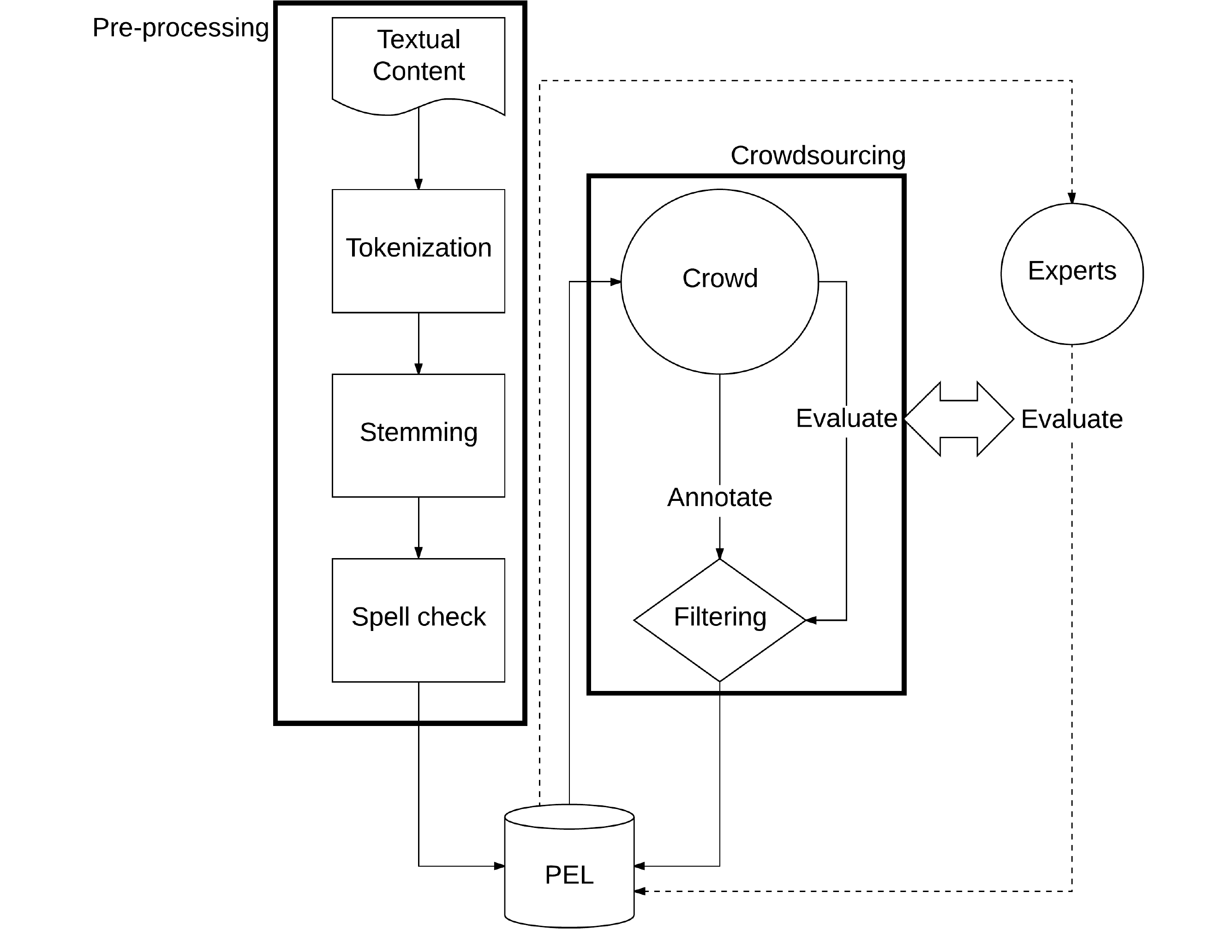}
\caption{PEL creation work-flow}
\label{fig:capture}
\end{figure}

Our aim is to create an end to end automated work-flow for the creation, evaluation and enrichment of a pure emotion lexicon. The work-flow, Figure~\ref{fig:capture}, can be separated in two main components. Pre-processing is the unsupervised process by which we derive the lexicon terms from any textual resource, while crowdsourcing deals with the crowdsourcing aspect of the lexicon. The Pure Emotions Lexicon includes emotional term groups, intensifiers and negators, and stop words.

Pre-processing is comprised of 3 unsupervised steps, tokenization, stemming and spell check. Textual content is tokenized as uni-grams, stemmed based on their rooted and checked for spelling. The resulting stems along with their stem groups are stored in a lexicon database. Crowdsourcing is using the lexicon database and the crowd to annotate each entry in the database. Participants submit their answers that go through a filtering process. If the answers are considered valid, they update the lexicon entries. The crowd also evaluates existing annotations, to determine the lexicon quality. As crowd evaluation methods are new in lexicon acquisition tasks, we compare crowd evaluations to those of expert linguists.

\subsection{Data}

During January 2017, we performed a keyword based crawl for articles and comments in the Europe subreddit\footnote{https://www.reddit.com/r/europe/} and tweets in Twitter, which contained the word "Brexit". The use of a political and controversial term in the query is deliberate, to capture the emotional diversity of a politically oriented corpus.

\subsection{Crowdsourcing}

The crowdsourcing task, hosted in CrowdFlower\footnote{\label{crowdflower}https://www.crowdflower.com/}, required contributors to label term groups in three different main classes, \textit{emotion}, \textit{intensifier} and \textit{none}, without a golden standard, rules or any participation restrictions . Emotion labelling included the 8 basic emotions as defined by Plutchik. Intensifier class included intensifiers and negators. Finally, \textit{none} referred to stop-words or words with no particular emotion.

Each of the eleven options for the main classes, will be referred to as "subclass". Terms are grouped based on their stem. Each term group has a main annotation class defined by majority, and several sub annotation classes, defined by the non majority annotations. However, to aid multi class analysis of the results, every annotation is logged in the lexicon database.

\begin{figure}[ht]
\includegraphics[width=1\textwidth]{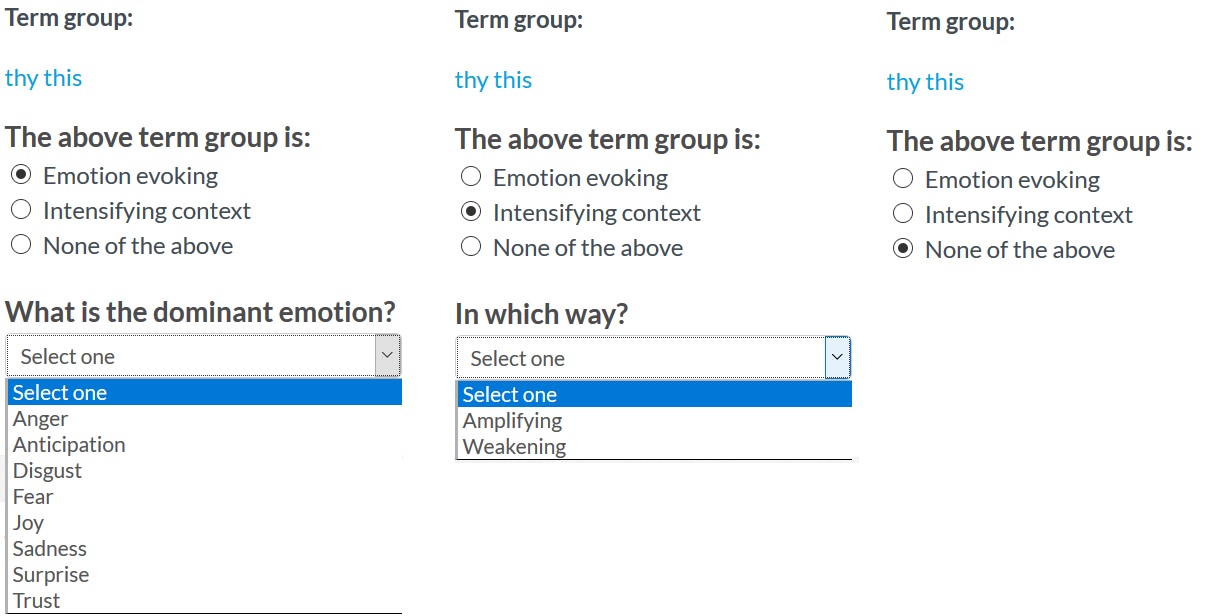}
\caption{Subclasses based on class selection}
\label{fig:subclasses}
\end{figure}

\subsection{Task Interface}

The task interface was the result of several experiments. Three major changes implemented based on these experimental interfaces were: the simplification of the task question, Figure~\ref{fig:subclasses}, the inclusion of only three main classes, and the replacement of words \textit{positive} and \textit{negative}, with \textit{amplifying} and \textit{weakening} in the intensifying class options. All experiments and the annotation task required highly experienced contributors, as defined by Crowdflower platform.

As seen in Figure~\ref{fig:subclasses} contributors select one of the three choices. If they choose \textit{Emotion evoking}, they are presented with a drop-down menu to choose from the eight basic emotions. Similarly, if they select \textit{Intensifying context} they have to specify whether it was \textit{Amplifying} or \textit{Weakening} the context, essentially annotating the intensifiers and negators. Finally if they choose \textit{None} they are presented with the next term group. To assist contributors with term definitions, every term group had a hyperlink to an English dictionary.

\subsection{Crowd}

More than one hundred eighty contributors performed eighty thousand annotations. By design, each user could not perform more than 660 unique annotations, excluding the assessment questions, to engage at least 100 contributors. Most of the workers annotated the maximum allowed terms, the first half of workers annotated 15\% of the term groups in our corpus, while the second half of workers annotate the rest 85\%. The simplicity of the task resulted in high overall worker engagement, with mean and median annotations per worker, at 429 and 580 respectively.

\subsection{Assessment}

Based on a set of experiments, we identified 136 term groups that would test the ability of a contributor in all of the three main classes, \textit{emotion evoking}, \textit{intensifying context}, and \textit{none}. As the assessment term groups had more than ten thousand annotations, we analyse it separately from the lexicon. 

In order for a worker to gain the ability to contribute to the crowdsourcing task and eventually get paid, he/she had to properly annotate 80\% of the assessment term groups encountered. The annotations should be within the dominant classes, and not subclasses, as defined from the assessment annotators. E.g., for an assessment term group that received 100 annotations in various emotions, we check if the worker annotates the term group as emotion evoking.

Let $W$ be the set of workers $w$ and and $S$ the set of eleven $s$ subclasses: eight emotions , two intensifiers, and none of the former. We define $s_{j_w} \;with\; j\epsilon\mathbb{Z}$ as the number of total annotations for each worker $w$. Then:

\begin{equation}\label{eq1}
\mu_w=\dfrac{max(s_{j_w})}{\sum_{j=1}^{11}s_{j_w}}
\end{equation}

We define $W_\alpha$ be the set of workers $w$ in the assessment process, $W_\beta$ the set of workers $w$ in the acquisition process. Then, for $x\epsilon\mathbb{Z} \;\&\;  x\epsilon(1,10)$ we define:

\begin{equation}\label{eq2}
W_{\alpha_x} = \{length\;of\;W_\alpha , \;w \ni \mu_w < x/10\}
\end{equation}

\begin{equation}\label{eq3}
W_{\beta_x} = \{length\;of\;W_\beta , \;w \ni \mu_w < x/10\}
\end{equation}

and:

\begin{equation}\label{eq4}
f_\alpha(x)= 1- \dfrac{W_{\alpha_x}}{W_\alpha}
\end{equation}

\begin{equation}\label{eq5}
f_\beta(x)= 1- \dfrac{W_{\beta_x}}{W_\beta}
\end{equation}

The optimal $\mu_w$ is found for $x \ni |f_\alpha(x)-f_\beta(x)|=min(|f_\alpha-f_\beta|)$. For this study, the optimal filtering percentage was found at 40\%, $x=4$.

\section{Evaluation}

\subsection{Data}

We crawled one hundred articles from Reddit, with more than forty thousand comments and more than three thousand tweets. For the Reddit  data, we collected information on location, time, and the number of upvotes\footnote{Denotes approval and/or agreement in Reddit}. For the Twitter data, we stored the number or re-tweets and favourites\footnote{Denote approval and/or agreement in Twitter}, time and location information.

Our focus is on single term (also known as unigram) sentiment, thus posts in both networks were processed to a single term list. In total, the number of unique terms in our corpus was 30227. Based on the enchant python library\footnote{https://pypi.python.org/pypi/pyenchant/}, used in \cite{cabot2016sibm}, and the supported Great British English dictionary, 19193 were validated and 11034 were invalidated. Our analysis will focus on the 19193 valid terms, that follow Zipf's Law with scaling-law coefficient $a=1$ is a good fit.

After validation, terms were stemmed with Porter Stemming Algorithm \cite{van1980new}. Stemming identified 10953 distinct term groups with one or more terms. Stop-words, intensifiers and negators are also included in the valid term groups. Both term validation and stemming are unsupervised, since our goal is to maintain scalability in  beyond polarity lexicon acquisition and sentiment classification.

\subsection{Assessment}

Workers from India and Venezuela, who contributed 92\% of the task, have annotated more than 30\% of the term groups with \textit{joy}. However, annotations from countries with more than 300 annotations, don't follow the same distribution. Specifically, workers from Philippines, United States, Colombia, Poland, United Kingdom, Russia, and Egypt, performed a smoother distributed emotion annotation. In comparison, the most annotated emotion in \cite{mohammad2013nrc} was fear in 18\% of the total terms.

By further analysing worker annotation distribution, we identified workers that had a significant fraction of their total annotations in a single subclass. E.g. one specific worker annotated 99\% of the assessment term groups he encountered as \textit{joy}. Dishonesty or spamming is a known problem in crowdsourcing \cite{difallah2012mechanical} and multiple proposed solutions exist\cite{difallah2012mechanical}, but they require gold standards or objective crowdsourcing tasks.

As we don't have a gold standard, and the task is more subjective, these spamming elimination methods are not applicable. Our solution is the implementation of a fast and efficient filter, which only relies on the obtained annotations and the assessment. If workers' answers were above a certain percentage on a single subclass for both the assessment and the annotation process, then the user would be flagged as dishonest and the total of their annotations would be discarded. This rule was applied to the whole range of possible answers, including the 8 emotions, 2 intensifiers and "none".

\begin{figure}[ht]
\includegraphics[width=1\textwidth]{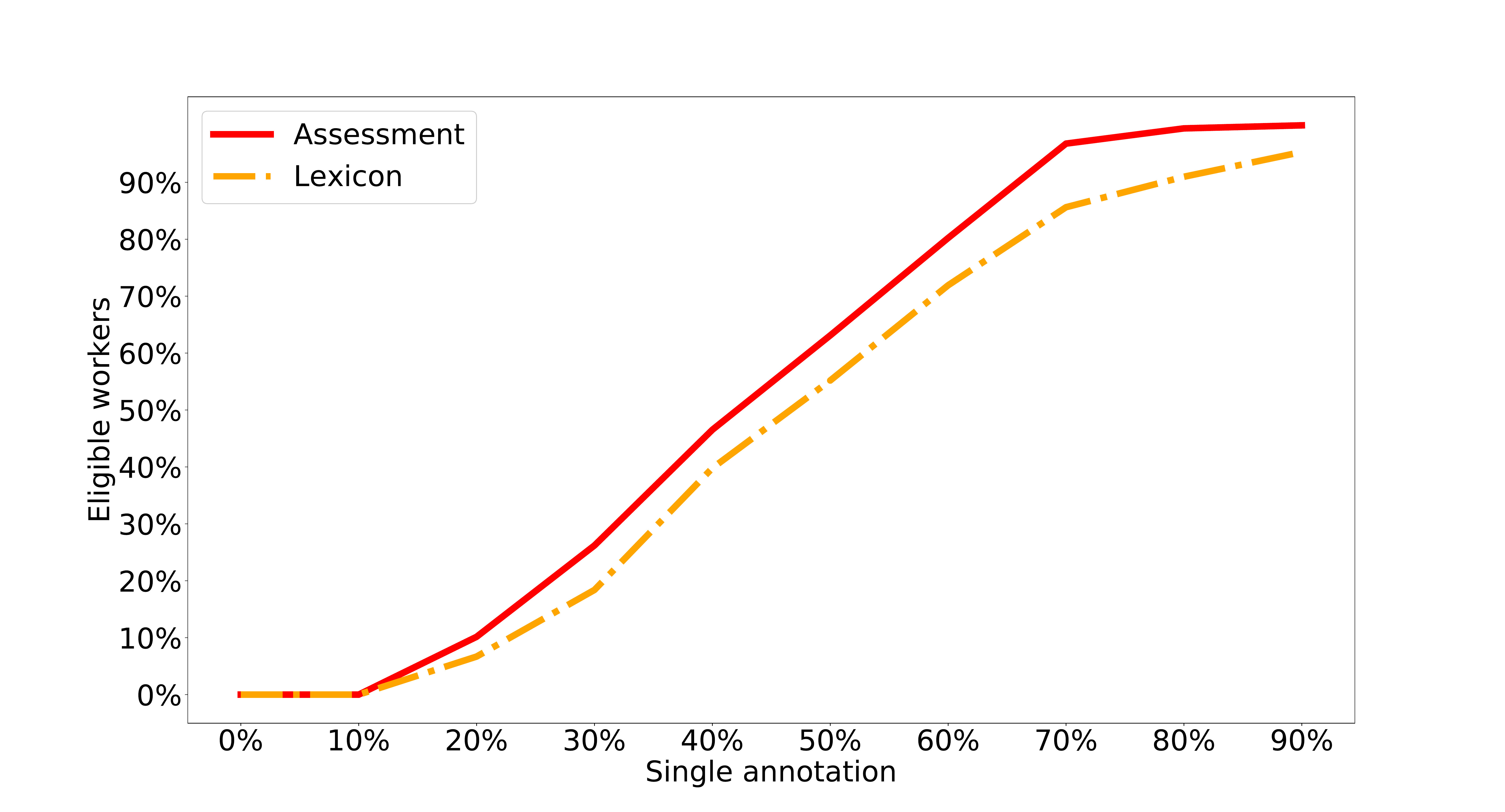}
\caption{Percentages of excluded workers over single annotation}
\label{fig:spamfilter}
\end{figure}

Prior to implementing the filter, we analysed how it would affect the number of eligible, not considered as spamming, workers. The thick line in Figure~\ref{fig:spamfilter} shows the percentage during the assessment term groups, and the dotted line shows the percentage during the lexicon acquisition.

The higher the single annotation percentage, the higher the certainty of spamming behaviour. In large part, the exclusion rate was similar for both the assessment and the lexicon annotations. A number of workers had a more cautious behaviour in the test questions, resulting in reduced percentage of exclusions during assessment process. This behaviour is justified, as the first set of questions encounter by a worker are knowingly a set of assessment questions.

Each assessment term group was annotated more than 120 times, by 187 annotators. This is a critical mass of contributors and provides valuable findings with regards to task. These are:

\begin{itemize}
  \item Workers rarely clicked the informative dictionary link. As a result, they would annotate emotional words to \textit{none}, probably due to misinterpretation. We avoided the direct inclusion of the dictionary definition, as it could be considered as a form of leverage. E.g. "vehement, halcyon" , two -uncommon- emotion baring words, were both annotated as \textit{none}, and less than 0.2\% of the workers (on average) clicked a dictionary link.
  \item The concept of intensifiers is understandable but requires critical mass \cite{markus1987toward}. A small number of annotators would initially annotate intensifiers/negators with an emotion, but the distribution would slowly shift towards the correct class. E.g. "reduce, little, plethora" , were initially annotated as \textit{sad} \textit{sad} \textit{joy}, but after tens of annotations they were annotated as \textit{weakening} \textit{weakening} \textit{intensifying}.
  \item All words should be evaluated, even those that seemingly don't carry a specific emotion. As times change, words and emotions acquire new links. E.g. "anti, serious" , were both annotated as fear evoking with a great emotional diversity.
\end{itemize}

\subsection{Lexicon Analysis}

The lexicon (will be referred as simply	 "PEL") is created after the exclusion of annotations following the 40\% single annotation filtering check.  We received more than seventy thousands annotations for 10593 term groups, of those only 22 thousand annotations for 9737 term groups are included in the final lexicon, as a result of filtering. Each term group had a mean 2.3 annotations from a total of 95 different annotators. Although the number of mean annotations in lexicon is less than half the mean annotations in the unfiltered corpus, the PEL annotations are considered of higher quality.

\begin{table}[ht]
\centering
\caption{Sample of non-emotional annotated term groups}
\label{table:subclass}
\begin{tabular}{ccc}
\textbf{Intensifiers}  & \textbf{Negators} & \textbf{None}\\
harder&dispensation dispense &is\\
largely large&minimize minimal&because\\
mostly&eliminates eliminated&to\\
\end{tabular}
\end{table}

Each lexicon term group has multiple subclass annotations, and the main subclass is defined by majority.  Even after filtering, the dominant emotion in our lexicon is \textit{joy}, while the least annotated emotion is \textit{disgust}. Additionally, 148 terms were annotated as intensifiers, 43 terms as negators, and 6801 terms as \textit{none}. A sample of five terms for each of the three subclasses can be seen in Table~\ref{table:subclass}. The full lexicon can be found on github\footnote{https://github.com/GiannisH/Lexicon/blob/master/lexicon.csv}.

Intensifiers and negators serve as modifiers to the emotional context of a word. Workers identified mostly valid intensifiers and negators that can modify emotion evoking words, in the absence of context. Judging from the received annotations, there is room for improvement on the description of the intensifier class and the provided examples, as a number of non intensifying words were falsely annotated.

Terms in our lexicon are grouped based on their stem. Stemming significantly reduced cost (by half) and time-required for the task. Grouping terms may create unnecessary multi-class annotations agreements, for terms in the same term group that might have different meanings. Annotation agreement refers to equal number of annotations in multiple subclasses or emotions. However, the vast majority of term groups in our lexicon, don't display any form of contradicting annotation. Contradicting emotions are portrayed in opposite edges of the circumplex Figure~\ref{fig:circumplex}, while emotional combinations are decribed in \cite{plutchik1980general}. In the lexicon, only 21\% and 20\% of the term groups had a subclass and an emotional agreement respectively. With regards to emotion, contradicting or multi-emotion agreement, could be observed in only 8.6\% of the total term groups.

Let $t$ be a term group in the lexicon and $S$ the set of eleven $s$ subclasses: eight emotions , two intensifiers, and none of the former. We define $s_j \;with\; j\epsilon\mathbb{Z}$ as the number of annotations for each term group $t$. For each $t$, the annotations for emotion subclasses are $s_j, j\epsilon[1,8]$, the annotations for intensifying subclasses are $s_j, j\epsilon[9,10]$, and the number of \textit{none} annotations is $s_j, j=11$.

Therefore, each $t$ can have an monotonically increasing finite sequence $a_{t}=a_1,....,a_n$ with $n=11$, where:
 
\begin{equation}\label{eq:6}
a_1=min(s_j) \leq . . . . \leq a_n=max(s_j)\\[10pt]
\end{equation} 

We say that term group $t$ has subclass agreement if and only if:
	
\begin{equation}\label{eq:7}
a_n=....=a_{n-k}, \; k\epsilon[1,10]\\[10pt]
\end{equation}

While $t$ has emotional agreement if and only if there is a subclass agreement with the sequence $A_t=a_n,....,a_{n-k}$ and:
\begin{equation}\label{eq:8}
\forall a_i \epsilon A_t\; where \; a_i=s_l, \;\;l\epsilon\mathbb{Z},\;l\epsilon[1,8]\\[10pt]
\end{equation}

Subclass agreement, refers to equal annotations, between emotional subclass(es) and at least one non-emotional subclass, or between multiple non-emotional subclass, Equation~\ref{eq:7}. On the other hand, emotional agreement refers to multiple emotion subclasses with equal annotations, Equation~\ref{eq:8}.

The number of subclasses in agreement and the number of terms in a term group are negatively correlated. Term groups with two terms appear to have the highest subclass agreement with exactly two subclasses. The most common occurring agreements are subclass \textit{none} paired with an emotion, and \textit{joy} paired with an emotion. The number of multi-class agreement occurrences is disproportional to the number of terms in a term group. This is a strong indication that stemming didn't confuse workers.

Similarly, for emotional agreement, the number of occurrences is disproportionate to the number of terms in the term group. Furthermore, emotional agreement appeared in 10\% of the term groups, while subclass agreement was found in 20\% of the term groups. In the agreement annotations, \textit{joy} is the most common emotion.  According to Plutchik's circumplex Figure~\ref{fig:circumplex}, each emotion has a contradicting one, and pairs of emotions indicate a more "complex" emotion. There are 697 emotional agreeing term groups, of 1434 terms, with exactly two emotions. These emotional dyads\cite{plutchik1980general} can be combined as seen in Table~\ref{table:dyads}. Simple basic emotion annotation tasks can indirectly provide complex emotional annotations. 

\begin{table}[ht]
\centering
\caption{Sample of combination dyads}
\label{table:dyads}
\begin{tabular}{cccc}
\textbf{Dyad}  & \textbf{Emotion} & \textbf{Term groups}& \textbf{Terms}\\

trust joy&love&94&231\\
joy anticipation&optimism&58&142\\
surprise joy&delight&43&88\\
fear joy&guilt&39&89\\
\end{tabular}
\end{table}

Dyadic emotional agreements could be interpreted as the resulting complex emotion, or further annotated to obtain a single dominant emotion.  There was a number of term groups with opposite emotion dyads, presented in Table~\ref{table:opposites},but as the number of annotations increases, emotional agreement occurrences -combination or opposition- decreases.

\begin{table}[ht]
\centering
\caption{Opposition dyads}
\label{table:opposites}
\begin{tabular}{ccc}
\textbf{Dyad} & \textbf{Term groups}& \textbf{Terms}\\
sadness joy&55&90\\
anger fear&20&34\\
surprise anticipation&16&30\\
disgust trust&12&18
\end{tabular}
\end{table}

In total, the lexicon features 17740 annotated terms with 3 classes and 11 subclasses.The dominant class for 7030 terms was \textit{emotion},  191 \textit{intensifying}, 6801 \textit{none}, and 3718 in some form of subclass agreement. Lexicon terms are mainly \textit{joy} annotated, and emotional agreement is prevalent in 10\% of the terms. Only 21\% of total terms have a subclass agreement.

\subsection{Reliability}

Single annotation reliability agreement is the degree of agreement between annotators, for term groups that have annotation majority in exactly one sub  class. In our lexicon, single annotation reliability agreement was low, mainly due to the low number of annotators for each term group in relation to the high number of possible categories.

Based on Fleiss Kappa \cite{fleiss1971measuring} (simply referred as \textit{k}), and as seen in Table~\ref{table:fleiss1}, term groups with 2 annotations had the lowest reliability agreement, while term groups with 6 annotations the highest reliability agreement. As the number of annotators rises, the number of possible agreement permutations increases but the number of major annotated subclasses decreases. More annotators have a positive effect in both \textit{k} and certainty of classification.

\begin{table}[ht]
\centering
\caption{Fleiss \textit{k} for different numbers of total annotations}
\label{table:fleiss1}
\begin{tabular}{ccc}
\textbf{Total annotations}  & \textbf{Subclass \textit{k}}& \textbf{Emotional \textit{k}}\\
2&0.143&0.122\\
3&0.135&0.187\\
4&0.161&0.171\\
5&0.184&0.244\\
6&0.188&0.252\\
\end{tabular}
\end{table}

As we restrict our lexicon to emotions, reliability increases for any number of annotators except two. This is explained by the decrease in the number of possible categories. When we restrict our analysis on emotion related annotation the probability for agreement in annotations increases, resulting in a high emotional \textit{k}. The best way to increase \textit{k} is to provide additional annotations that will eventually converge to a majority class or a limited group of classes.

\begin{figure*}[ht]
\includegraphics[width=1\textwidth]{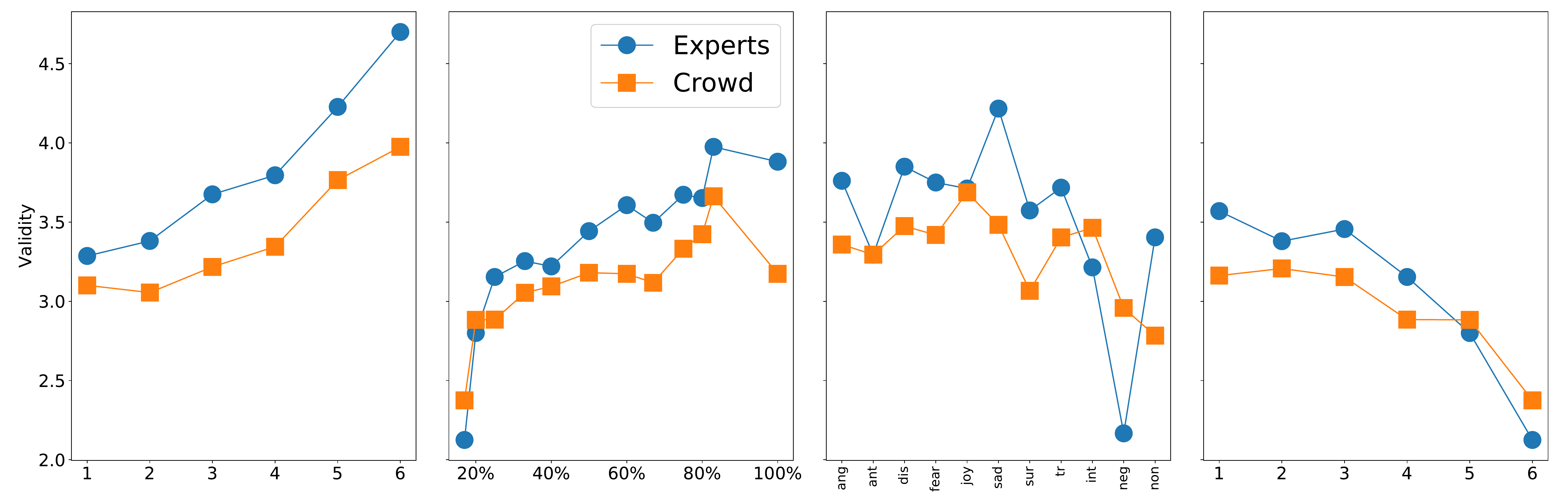}
\caption{Validity over: Majority subclass annotations, Majority subclass annotations percentage, Subclasses, Subclass Agreement}
\label{fig:celexicon}
\end{figure*}

\subsection{Crowd and experts comparison}

\subsubsection{Lexicon}

We perform a direct comparison of expert and crowd contributors, for 1000 term groups based on the number of total annotations(200 term groups with 2 total annotations, 200 term groups with 3 total annotations, and so on up to term groups with 6 total annotations). The experts are two Ph.D. linguists, while the crowd is made up of random high quality contributors that choose to participate in the task. As a reference, the cost of hiring two experts is equal to the cost of employing nineteen contributors in Crowdflower.

Evaluators were given a summary of the annotations received for the term group in the form of:\textit{The term group "inequality inequity" received annotations as 50.0\% sadness, 33.33\% disgust, 16.67\% anger.} Then, they were asked to evaluate on a scale from 1 to 5, how valid these annotations were considered.

The summary of the evaluation for both experts and crowd can be seen in Figure~\ref{fig:celexicon}. The first graph presents the validity over the number of annotations in the main class of the term group. Although this information is hidden from the evaluators, high annotational agreement results in high evaluation scores. Both experts and the crowd follow that positive trend. Crowd contributors are more strict in their evaluations, but after four annotations we observe a significant validity increase on both crowd and experts.

Likewise, the annotation percentage for the majority class has a positive influence to the evaluation score, with the exception of 100\% agreement, second graph Figure~\ref{fig:celexicon}. The weighing factor for term groups with 100\% annotation agreement is the reduced number of total annotations, as the mean number of total annotations drops abruptly on the 100\%, and total agreement is more frequent in term groups with low number of total annotations. It's worth noting that certain percentages can only occur on specific number of total annotations, e.g. 17\% and 83\% can only occur when the number of total annotations is six.

In emotion annotations, as seen on the third graph of Figure~\ref{fig:celexicon} crowd and experts follow a similar evaluation pattern. \textit{Anticipation} and \textit{joy} had the exact same evaluation, while every other emotion and stop words were evaluated lower from the crowd. The only subclasses evaluated higher from the crowd were intensifiers and negators, with a significant difference in the evaluations for the latter. Section 6.3 provides a more detailed evaluation for term groups that received at least one annotation as intensifiers or negators.

The final graph in Figure~\ref{fig:celexicon} presents a clear negative correlation of subclass agreement and evaluation scores. The highest number of subclasses that do not affect  evaluation scores is three, above that there is a steady decline of the evaluation scores, for both the crowd and the experts.

The evaluation results provide some key insights in the importance of the number of annotations. The evaluation scores start to improve after four annotations. Annotational agreement and majority voting are less important. Subclass agreement has a negative effect on three or more subclasses. Most importantly and compared to experts, the crowd is a stricter evaluator with significantly lower costs, and higher scalability. Since strict evaluation leads to higher quality annotations, the evaluation can be performed by the crowd instead of experts. Crowd contributors can be found in high numbers and multiple platforms, compared to expert linguists.
\subsubsection{Intensifiers and negators}

\begin{figure}[ht]
\includegraphics[width=1\textwidth]{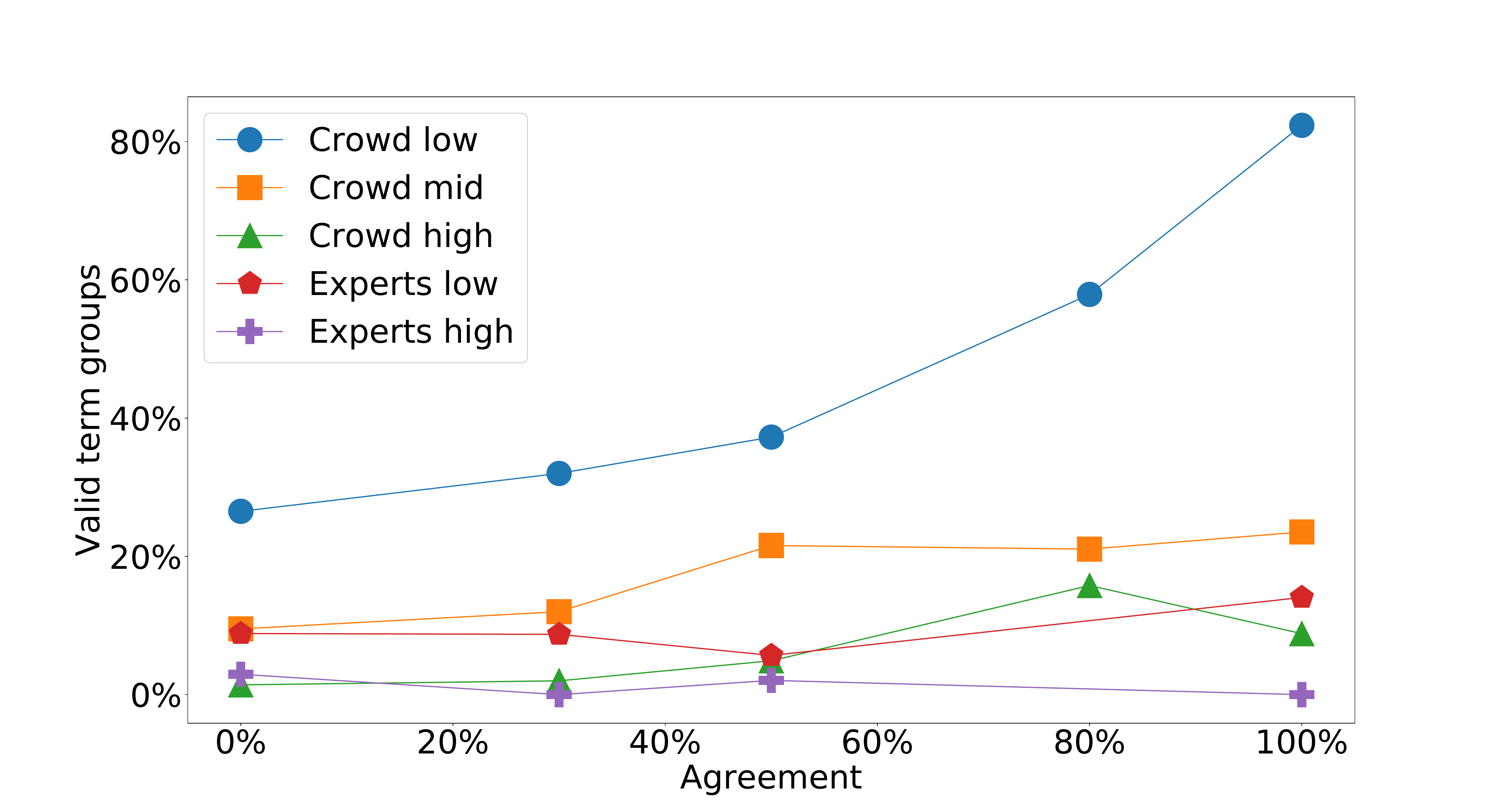}
\caption{Intensifying class evaluations}
\label{fig:ceintensifiers}
\end{figure}

Evaluation of intensifiers and negators, was also a batch of evaluation and annotation tasks, as mentioned in Section 6.2. However, the difference was that now evaluators had to answer if a term group included at least one valid intensifier or negator. The evaluation was again performed by experts and the crowd, as described in Section 6.2.1. Based on the annotations received in PEL, we used 541 term groups that had at least one annotation in any of the intensifying subclasses. Although, the particular selection of term groups is statistically significant, we expect relatively low evaluation scores. That is because the number of intensifying annotations is low in most of the selected term groups.

In Figure~\ref{fig:ceintensifiers}, we define varying levels of agreement on the validity of the intensifying class, based on the agreement of evaluations. For the experts group, \textit{low agreement} refers to term groups that received at least one out of two evaluations as valid, while \textit{high agreement} requires the evaluation agreement of both experts. Similarly for the crowd, \textit{low agreement} refers to a minimum of two valid evaluations, \textit{mid agreement} corresponds to at least three, and \textit{high agreement} requires an absolute agreement of all four evaluators. 

Experts are far more strict than the crowd in the evaluation of intensifiers and negators. When the validity agreement is low on both evaluation groups, the average valid term group difference is more than 40\%, but the high validity agreement the difference is just 5.33\%. When high agreement evaluation is applied, the crowd and expert evaluations are almost identical. The number of crowd evaluations is the factor that provides a degree of freedom in the evaluation strictness.

\section{Limitations}

Lexicon acquisition is a complex task that includes a mixture of objective and subjective tasks. While annotation of emotions is more subjective, annotation of linguistic elements (such as stop words, emotion shift terms, intensifiers etc.) is purely objective. We presented a novel work flow that provides quality results for both subjective and objective tasks.

Subcomponents of the lexicon acquisition could be improved on an individual basis. Spell check can include spelling recommendations, filtering could incorporate rewarding and penalties, evaluation process can include experts and so on.

Crowd diversity in the annotation and evaluation process is another limiting factor. Ideally we would prefer a fixed number of individuals, each to annotate and evaluate the whole corpus. However, the uniformity of expert judgement is replaced with the diversity and mass of contributors.

The corpus may be limiting the term groups in the lexicon to specific domain-specific subjects. Comparisons with existing lexicons, such as NRC\citep{mohammad2013nrc} indicate a moderate overlap with 40\% common terms. Additionally, the number of annotations for a number of term groups is relatively low. However, the batch task of evaluation and annotation provided almost ten thousand annotations, and increased the mean number of annotations from 2.3 to 3.2.

\section{Conclusion and future work}

We demonstrated that the crowd is capable of producing and evaluating a quality pure emotion lexicon without gold standards. Our work-flow is unsupervised, significantly lower costs, and improves scalability. There are however, various parameters that should be taken into account. Spam is very common and quality assessment post-annotations should be implemented.

Our approach required workers to label term groups as emotion, intensifiers, and stop words. Agreement is not necessary and multi emotional term groups, with up to three emotions, are considered equally valid to single emotion term groups. The hardest task for the crowd proved to be the classification of intensifiers and negators, probably because it required a certain level of objectivity which contradicted the overall subjectivity of the emotional annotation task. Based on the evaluation of term groups and the results from the assessment, as the number of overall annotators rises the number of valid annotations increases proportionally. This indicates the importance of a critical mass in lexicon acquisition tasks.

Stemming reduced time and costs requirements, with minimal emotional and subclass agreement. Costs were reduced by 45\%, and multi-emotion classification was lower than 10\%. Term groups did not create confusion amongst workers, and only a small fraction of term groups had subclass agreement. On the contrary, including the stem and description in the task confused workers, and were excluded from the interface. We tested several interface designs, and the one that worked best had minimal instructions. Lexicon acquisition interfaces in paid micro-task environments should be further studied, with regards to various other contribution incentives.

The crowd is as capable of evaluating lexicons, as experts. Linguistic element evaluation can be efficiently crowdsourced, and the evaluation of emotional or non emotional elements can be as strict as needed. The number of evaluators plays a key role in both emotional and linguistic evaluations. The crowd is strict on emotional evaluations, while the experts are strict in linguistic evaluations. However, a high number of crowd evaluations broadens the strictness freedom, with a small fraction of the experts' hiring costs. Depending on the number of evaluations, varying levels of evaluation agreement can be implemented.

Our long term goal is to create a voluntary platform for pure emotion lexicon acquisition, to further study the effects of critical mass in lexicon acquisition. In short term, we will perform the exact same crowdsourcing task in a voluntary platform, Crowd4U\footnote{https://crowd4u.org/en/} or similar platforms, to study the effect of monetary and contribution incentives in pure emotion sentiment annotation. In parallel, we will perform a qualitative analysis with regards to understanding of intensifiers and negators, to create the optimal set of instructions and examples. Finally, we are considering how we can extend the approach to various other linguistic elements, such as words that split the sentence, words that indicate more important parts of a sentence and so on.

We believe that beyond polarity sentiment analysis can enhance and extend simple polarity based applications. Sentiment analysis in marketing, politics, health monitoring, online social networks, and evaluation processes would benefit from a crowdsourced pure emotion lexicon.

\end{sloppypar}

\section*{Acknowledgement}

This work was supported by the EU project "QROWD - Because Big Data Integration is Humanly Possible".

\bibliographystyle{unsrt}
\bibliography{paper3} 

\end{document}